\documentclass{bmvc2k}
\usepackage[toc,page]{appendix}

\title{Tensor Component Analysis for Interpreting the Latent Space of GANs}

\addauthor{James Oldfield}{j.a.oldfield@qmul.ac.uk}{1}
\addauthor{Markos Georgopoulos}{m.georgopoulos@imperial.ac.uk}{2}
\addauthor{Yannis Panagakis}{yannisp@di.uoa.gr}{3}
\addauthor{Mihalis A. Nicolaou}{m.nicolaou@cyi.ac.cy}{4}
\addauthor{Ioannis Patras}{i.patras@qmul.ac.uk}{1}

\addinstitution{
 Queen Mary University of London, UK
}
\addinstitution{
 Imperial College London, UK
}
\addinstitution{
    University of Athens, Greece
}
\addinstitution{
    The Cyprus Institute, Cyprus
}

\runninghead{Oldfield et al.}{TCA for Interpreting the Latent Space of GANs}

\usepackage[capitalise]{cleveref}
\usepackage{hyperref}
\usepackage{amsmath}
\usepackage{amssymb}
\usepackage{dsfont}
\usepackage{algorithm}
\usepackage{algpseudocode}

\usepackage{svg}
\usepackage{multirow}

\usepackage{graphicx}

\usepackage{caption}

\usepackage{svg}
\newcommand{\brows}[1]{%
  \begin{bmatrix}
  \begin{array}{@{\protect\rotvert\;}c@{\;\protect\rotvert}}
  #1
  \end{array}
  \end{bmatrix}
}
\newcommand{\rotvert}{\rotatebox[origin=c]{90}{$\vert$}}
\newcommand{\rowsvdots}{\multicolumn{1}{@{}c@{}}{\vdots}}

\usepackage{amsthm}
\theoremstyle{definition}

\newtheorem{proposition}{Proposition}

\definecolor{ForestGreen}{RGB}{34,139,34}

\providecommand{\realnum}{\mathbb{R}}

\begin{document}
\maketitle
\begin{abstract}
    This paper addresses the problem of finding interpretable directions in the latent space of pre-trained Generative Adversarial Networks (GANs) to facilitate controllable image synthesis. Such interpretable directions correspond to transformations that can affect both the style and geometry of the synthetic images. However, existing approaches that utilise linear techniques to find these transformations often fail to provide an intuitive way to separate these two sources of variation. To address this, we propose to a) perform a \textit{multilinear} decomposition of the tensor of intermediate representations, and b) use a tensor-based regression to map directions found using this decomposition to the latent space. Our scheme allows for both linear edits corresponding to the individual modes of the tensor, and non-linear ones that model the multiplicative interactions between them. We show experimentally that we can utilise the former to better separate style- from geometry-based transformations, and the latter to generate an extended set of possible transformations in comparison to prior works. We demonstrate our approach's efficacy both quantitatively and qualitatively compared to the current state-of-the-art.
\end{abstract}

\section{Introduction}
\label{sec:intro}

\begin{table}[t]
\centering
\resizebox{\textwidth}{!}{%
\begin{tabular}{l|l|l|l}
\textbf{Transformation type} & \textbf{`Style'} & \textbf{`Geometry'} & \textbf{`Multilinear Mixing'} \\ \hline
\textbf{Examples} & illumination, colour, background & yaw, pitch, translation & deformation, skew, distort \\ \hline
\textbf{Prominent modes} & Channel mode & Spatial modes & Interactions between modes
\end{tabular}%
}
\caption{The hierarchy of transformations obtainable (and where) with our method.}
\label{tab:transformation-types}
\end{table}

Over the past few years, GANs \cite{goodfellow2014gan} have continued to push the state-of-the-art forward for the task of image synthesis. Many recent works have proposed more sophisticated architectures for improving the quality of the generated images: such as with the use of transposed convolutions \cite{radford2016dcgan}, progressive growing \cite{karras2018prog}, or with explicit modulation of the style content \cite{karras2019style}. As the quality of the images generated by these methods continues to improve, there is increasing interest in exploring how one can better control the image synthesis process.
A prominent research direction with this goal in mind is that of finding directions in the latent space that reliably affect interpretable modes of variation \cite{goetschalckx2019ganalyze,gansteerability} in the generated images \cite{plumerault20iclr,hark2020ganspace,voynov2020unsupervised,shen2020interfacegan,shen2020interpreting,shen2021closedform,Tzelepis_2021_ICCV}. Prior works showed that with supervision, one can isolate factors such as age, gender, and pose \cite{shen2020interfacegan,shen2020interpreting}--facilitating the ability to modify these attributes in an image by a desired amount.
Despite this success, the necessary supervision requires expensive manual labour, highlighting the need for an unsupervised discovery of such directions. In this vein, recent methods use auxiliary networks to search for `diverse' image transformations \cite{voynov2020unsupervised,Tzelepis_2021_ICCV}, decompose the weights defining the mapping between layers \cite{shen2021closedform}, or decompose the intermediate generator's representations directly \cite{hark2020ganspace}.
However, the latter's approach of treating this tensor of intermediate activations as a vector entangles the variation in both the spatial and channel modes.
With these approaches, there is often no intuitive way to separate different types of transformations. Such an ability has recently been shown to be useful for a number of downstream tasks, such as saliency detection \cite{voynov2020unsupervised} and unsupervised object segmentation \cite{voynov2020big}. To this end, we take inspiration from the categorisations introduced in \cite{voynov2020unsupervised} and focus on locating two different types of interpretable directions: \textit{style}-based directions (such as illumination and colour) and \textit{geometry}-based directions (such as orientation and translation). This categorisation is defined in \cref{tab:transformation-types}, along with examples of the types of transformations we seek to find.

Motivated by findings from the style transfer literature \cite{huang2017arbitrary}, we suggest in this work that the modes of the tensors of activations in a generator can be useful for isolating these different types of semantic transformations.
To address this, we propose a multilinear approach that finds interpretable directions in the latent activations' natural tensorial form $\mathcal{Z}\in\mathbb{R}^{C\times H \times W}$ as shown in \cref{fig:teaser}.
We learn a separate basis $\mathbf{U}^{(C)},\mathbf{U}^{(H)},\mathbf{U}^{(W)}$ for the channel, height, and width modes of the tensors, respectively, to locate the various types of transformations described above.
Such a multilinear treatment comes with a number of benefits beyond being computationally cheaper than its linear counterpart. Firstly, this leads to an implicit separation of style and geometry: this allows one to find interpretable directions corresponding semantically to each mode of the tensor.
What's more, we show how the multiplicative interactions \cite{jayakumar2019multiplicative} of basis vectors across the modes of the tensor (`multilinear mixing' in \cref{tab:transformation-types}) correspond to transformations unobtainable with the linear treatment (such as forehead shape, where both the width and height dimensions influence the attribute together in a non-uniform manner). We propose the use of a tensor regression to map these combinations of basis vectors in the form of a tensor back to latent space--with a low-rank structure to offer flexibility over the extent to which the interpretable directions are similar to the transformations found in the training data. We demonstrate with a series of experiments on multiple generator architectures and datasets the validity of our method--including showing superior disentanglement both qualitatively and quantitatively over prior work. Our main contributions can be summarised as follows:

\begin{figure}[t]
    \centering
    \includegraphics[width=\linewidth]{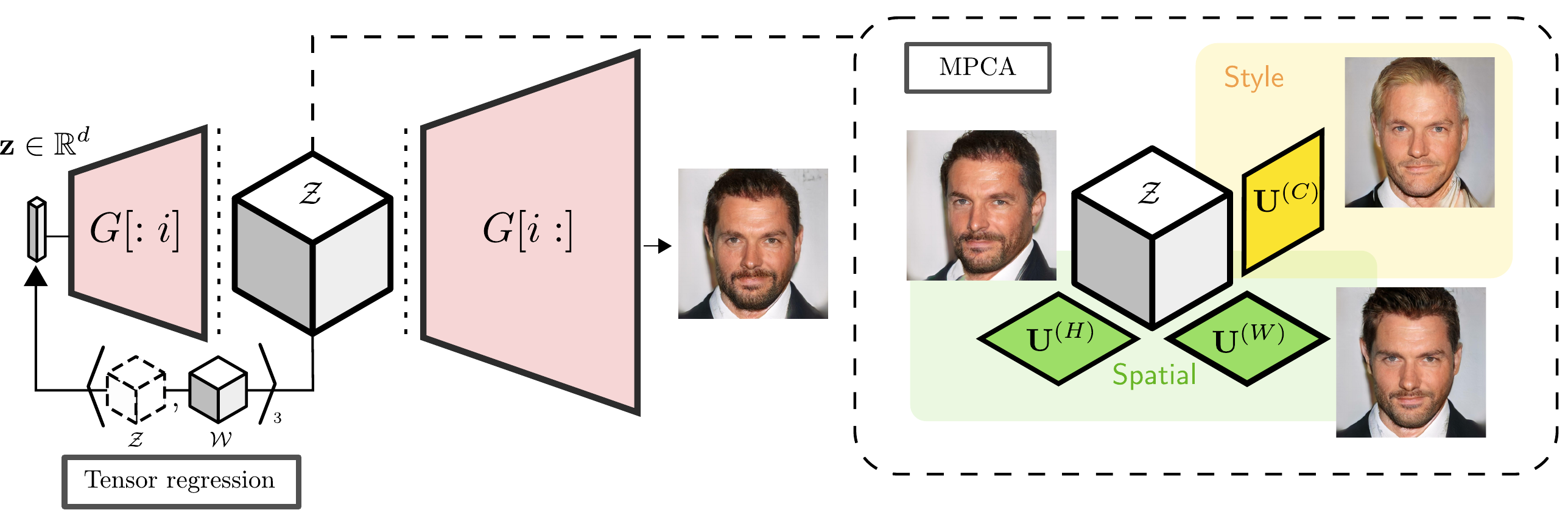}
    \caption{
    An overview of our proposed method: on a pre-trained generator's intermediate features $\mathcal{Z}\in\mathbb{R}^{C\times H \times W}$ we perform mutlilinear PCA (right) to learn a basis for each mode of the tensor--localising different types of interpretable directions to each mode. We learn a tensor regression from the activations back to the latent code, allowing us to then map combinations of the multilinear bases back to interpretable directions in latent space.}
    \label{fig:teaser}
\end{figure}

\begin{itemize}
   \setlength\itemsep{0em}
    \item We propose an intuitive way to separate different types of interpretable directions in a GAN by decomposing the activation maps in the generator in a multilinear fashion. We show how the linear approach of \cite{hark2020ganspace} can be framed as a special case.
    \item We show that by modelling the \textit{interactions} of basis vectors across modes, one can recover transformations that are the influence of multiple modes in a non-uniform fashion (such as thickening of the forehead), and are hence unobtainable with linear approaches.
    \item We propose the use of a tensor regression for mapping activations back to their latent code, which provides regularisation in the form of a low-rank structure on the weights.
    \item We demonstrate both qualitatively and quantitatively that prominent directions along each mode learnt with our method correspond to attributes such as hair colour, pitch, or yaw. These found directions showcase superior disentanglement over the state-of-the-art methods in terms of the style or geometry categories of transformations identified in \cref{tab:transformation-types}.
\end{itemize}

\section{Related work}
\label{sec:related-work}

\paragraph{GANs \& Interpretable directions}
The seminal work of GANs \cite{goodfellow2014gan} showed how, using adversarial training, one can train a so-called generator to map a latent noise vector to a high resolution synthetic image. This generator is often realised with a convolutional neural network \cite{radford2016dcgan, karras2018prog}, with various tricks having been developed to improve the quality of the samples \cite{karras2019style,Karras_2020_CVPR,brock2018large}. Interestingly, \cite{radford2016dcgan} showed that one can perform arithmetic in the latent space that affects predictable changes in image space. Since these works, a host of methods have been proposed to explore the latent structure in these generators by imposing structure at training-time \cite{infogan,highfid2020wonkwang} or more recently in the pre-trained generators themselves \cite{voynov2020unsupervised,shen2020interfacegan,shen2020interpreting,hark2020ganspace,broad2020bending,Tzelepis_2021_ICCV}. However, of the approaches that decompose the intermediate features directly (such as \cite{hark2020ganspace}), a linear decomposition is applied--where we argue a multilinear one can be more suitable in providing an ability to locate different categories of transformation.

\paragraph{Tensor methods for visual data}
The use of multilinear structures or decompositions to represent and interpret visual data has a rich history. For example, bilinear models have been used to separate style and content \cite{tenenbaum_separating_2000}, and multilinear ones for learning the structure of visual data more generally \cite{vasilescu_multilinear_2002,wang2017disentangling,wang_learning_2017}.
More recently, a popular approach is to combine these tensor methods with deep learning methodologies \cite{tensormethods2021panagakis}. For example, for modelling \cite{mitigating2021georgopoulos,multilinear2020georgopoulos} multiplicative interactions \cite{jayakumar2019multiplicative}, disentangling the modes of variation \cite{wang2019adversarial}, or for interpreting or decomposing the weights and operations of a deep neural network \cite{kossaifi2020factorized,bulat2020incremental,chrysos2020p}.

\section{Methodology}
\label{sec:methodology}
In this section, we describe our method in detail. We first provide an overview of the relevant notation and definitions in \cref{sec:notation}. We follow in \cref{sec:formulation} by formulating the problem of learning interpretable directions, including the PCA solution in \cref{sec:linear-solution}. In \cref{sec:m-learning}, we introduce the proposed multilinear approach (formulating GANSpace as a special case), and finally in \cref{sec:m-editing} we detail how we use these learnt directions to modify the latent code.

\subsection{Notation}
\label{sec:notation}
Firstly, we detail the notation we adopt throughout this paper, and provide definitions of the relevant operations\footnote{We refer readers to \cite{kolda_tensor_2009} for a more detailed overview.}.
We use uppercase (lowercase) boldface letters to denote matrices (vectors), e.g. $\mathbf{X}$ ($\mathbf{x}$), and calligraphic letters for tensors, e.g. $\mathcal{X}$. We use $\mathds{1}_d\in\mathbb{R}^d$ to denote the vector of ones from the main diagonal of a $d$-dimensional identity matrix--the subscript of which we sometimes omit for brevity. The \textbf{Kronecker product} of two matrices $\mathbf{X}\in\mathbb{R}^{I_1 \times I_2}$ and $\mathbf{Y}\in\mathbb{R}^{J_1 \times J_2}$ is defined as
\begin{equation*}
\mathbf{X}\otimes\mathbf{Y} = \begin{bmatrix} x_{11} \mathbf{Y} & \cdots & x_{1I_2}\mathbf{Y} \\ \vdots & \ddots & \vdots \\ x_{I_11} \mathbf{Y} & \cdots & x_{I_1 I_2} \mathbf{Y} \end{bmatrix}
\in \mathbb{R}^{I_1 \cdot J_1 \times I_2 \cdot J_2}.
\end{equation*}
We refer to each element of an $N^{th}$ order tensor $\mathcal{X}$ using $N$ indices, i.e., $\mathcal{X}(i_{1}, i_{2}, \ldots, i_{N}) \doteq x_{i_{1} i_{2} \ldots i_{N}}\in\mathbb{R}$. The \textbf{mode-$n$ fibers} of a tensor are the column vectors formed when fixing all but one of the indices (e.g. $\mathbf{x}_{:jk}$), and can be seen as a higher-order analogue of matrices' rows and columns.
 For a tensor $\mathcal{X} \in \realnum^{I_1 \times I_2 \times \cdots \times I_N}$, we can arrange its mode-$n$ fibers along the columns a matrix, giving us the notion of the \textbf{mode-$n$ unfolding} denoted as $\mathbf{X}_{(n)} \in \realnum^{I_{n}
 \times \bar{I}_{n}}$ with $\bar{I}_{n}= \prod_{t=1 \atop t  \neq n}^N I_t $. Lastly, the \textbf{mode-$n$ (matrix) product} of a tensor $\mathcal{X}\in\realnum^{I_1\times I_2\times \dots \times I_N}$ and a matrix $\mathbf{W}\in\realnum^{J\times I_n}$ is denoted by $\mathcal{X}\times_n \mathbf{W}\in\realnum^{I_1\times \cdots \times I_{n-1} \times J \times I_{n+1} \times \cdots \times I_N}$. Most usefully for this paper, the mode-$n$ matrix product can be expressed in terms of the unfolded tensors as
\begin{equation*}\label{eq:mode-n}
    \mathcal{Y} = \mathcal{X} \times_n \mathbf{W} \quad \Leftrightarrow \quad \mathbf{Y}_{(n)} = \mathbf{WX}_{(n)}.
\end{equation*}

\subsection{Problem formulation}
\label{sec:formulation}

A pretrained generator $G$ receives a low-dimensional vector latent code $\mathbf{z}\in\mathbb{R}^d$ and maps it to a synthetic image $G(\mathbf{z}) =\mathcal{X}\in\mathbb{R}^{C\times H \times W}$. $G$ is usually implemented as a series of convolutional layers, meaning that the image has a number of intermediate representations $\mathcal{Z}$ in the form of a collection of feature maps. In traditional generator architectures, the output image $\mathcal{X}$ is parameterised fully by its latent code $\mathbf{z}$. Therefore by modifying the latent code, one can affect changes in the resulting synthetic image. The goal of learning interpretable directions is that of finding a latent vector $\mathbf{z}'$ corresponding to a high-level attribute of interest (such as `hair color') in image space. One can then modify the original latent code by some amount $\alpha\in\mathbb{R}$ to generate the modified image $\mathcal{X}' = G \left( \mathbf{z} + \alpha\cdot\mathbf{z}' \right)$.

\subsubsection{The linear solution: GANSpace}
\label{sec:linear-solution}

Let $\left\{ \mathcal{Z}_1, \mathcal{Z}_2, \dots, \mathcal{Z}_M \right\}$ be a batch of $M$ feature maps from an intermediate layer in a pretrained generator, each of which is a third order tensor $\mathcal{Z}_m\in\mathbb{R}^{C \times H \times W}$ with channel, height, and width modes. The PCA-based method of GANSpace \cite{hark2020ganspace} finds interpretable directions in the traditional architectures by first vectorising these tensors $\text{vec}\left( \mathcal{Z}_m \right)$ and then learning a PCA basis $\mathbf{U}$ that admits the following decomposition
\begin{align}
    \text{vec}\left( \mathcal{Z}_m \right)
        &= \mathbf{U}\mathbf{U}^\top \text{vec}\left( \mathcal{Z}_m \right) \label{eq:pca} \\
        &= \text{vec}\left( \mathcal{Z}_m \right) \times_1 \mathbf{U}\mathbf{U}^\top
        \label{eq:pca-moden}.
\end{align}
This basis $\mathbf{U}$ is then regressed back to the latent space with linear transformation $\mathbf{W}$, the columns of which then contain the interpretable directions in the original latent space. One can then make edits with $\mathcal{X}' = G \left( \mathbf{z} + \mathbf{Ws} \right)$, where $\mathbf{s}$ is a `selector' vector that takes a linear combination of the desired columns of the basis.

\subsection{Learning multilinear directions}
\label{sec:m-learning}

Rather than living in a single $C\cdot H \cdot W$-dimensional vector space however, each activation tensor $\mathcal{Z}_m$ can be seen more naturally as belonging to a tensor space formed by the outer product of \textit{three} vector spaces $\mathbb{R}^C \bigotimes \mathbb{R}^H \bigotimes \mathbb{R}^W$.
We thus retain the tensorial structure and perform a \textit{multilinear} PCA \cite{haiping_lu_mpca_2008}, allowing us to write each of the $M$ tensors in the batch as
\begin{align}
    \label{eq:mpca}
    \mathcal{Z}_m = \mathcal{Z}_m \times_1 {\mathbf{U}^{(C)}}{\mathbf{U}^{(C)}}^\top \times_2 {\mathbf{U}^{(H)}}{\mathbf{U}^{(H)}}^\top \times_3 {\mathbf{U}^{(W)}}{\mathbf{U}^{(W)}}^\top,
\end{align}
where each $\mathbf{U}^{(n)}$ forms an orthonormal basis for the space spanned by all $M$ tensors' mode-$n$ fibers. From this it is clear that the GANSpace \cite{hark2020ganspace} approach as we formulate it using the mode-$1$ product in \cref{eq:pca-moden} is a special case of our proposed formulation in \cref{eq:mpca}--instead operating on the flattened tensor with a single factor matrix.
To further shed light on the connection between our formulation and the linear approach of GANSpace in \cref{eq:pca}, we can rewrite each activation tensor in \cref{eq:mpca} in its vectorised form \cite{kolda2006multilinear} as
\begin{equation}
\label{eq:mpca-vec}
    \text{vec}(\mathcal{Z}_m) = \left(
        {\mathbf{U}^{(W)} \mathbf{U}^{(W)}}^\top \otimes
        {\mathbf{U}^{(H)} \mathbf{U}^{(H)}}^\top \otimes
        {\mathbf{U}^{(C)} \mathbf{U}^{(C)}}^\top
    \right) \text{vec}(\mathcal{Z}_m),
\end{equation}
highlighting the crucial differences between the linear and multilinear methods: MPCA models the interactions of the columns of all three bases.

\subsubsection{Computing the multilinear basis}
\label{sec:computing}

The factor matrices ${\mathbf{U}^{(n)}}^\top$ in \cref{eq:mpca} project the input tensor to a tensor subspace where the total scatter is maximised \cite{lu2013subspacebook,lu2011subspacesurvey}--this objective is a higher-order analogue of that of regular PCA. We thus follow Section. III. B of Lu et al. \cite{haiping_lu_mpca_2008} and compute factor matrix $\mathbf{U}^{(n)}$ as the matrix of left-singular vectors from the SVD of the following mode-$n$ total scatter matrix
\begin{align}
  \mathbf{U}^{(n)} \boldsymbol\Sigma {\mathbf{V}^{(n)}}^\top = \sum_{m=1}^M \left(\mathbf{Z}_{m(n)} - \mathbf{\bar{Z}}_{(n)}\right) \left(\mathbf{Z}_{m(n)} - \mathbf{\bar{Z}}_{(n)}\right)^\top,
\end{align}
where $\mathbf{Z}_{m(n)}$ is the mode-$n$ unfolding of the $m^\text{th}$ GAN sample's activation tensor, and $\mathbf{\bar{Z}}_{(n)}$ is the mean of all mode-$n$ unfoldings.
The columns of each of these factor matrices contain the principal directions for each mode.
Pseudocode outlining this procedure in a pre-trained GAN can be found in the supplementary material.

\subsection{Editing with multilinear directions}
\label{sec:m-editing}

Rather than taking a linear combination of the basis vectors like in the linear setting of \cref{sec:linear-solution}, we propose to instead form what we call an `edit tensor', $\mathcal{Z}'\in\mathbb{R}^{C\times H \times W}$ as a combination of the basis vectors across the modes, either separately or together. These different types of combinations of the basis vectors produce the various transformations in \cref{tab:transformation-types}. A graphical illustration of this process can be found in the supplementary material.

\paragraph{Mode-wise edits}

Each column of $\mathbf{U}^{(n)}$ is a basis vector for the space spanned by the mode-$n$ fibers.
We perform a mode-$n$ edit by adding one or more of these basis vectors to all mode-$n$ fibers of $\mathcal{Z}'$. For example, to perform an edit using the $i^\text{th}$ channel basis vector we add $\alpha_i \cdot \mathbf{u}^{(C)}_i \circ \mathds{1}_H \circ \mathds{1}_W$ to the edit tensor--where $\alpha_i$ is a scalar controlling the relative weight. More generally, to add the $i^\text{th}$ basis vector for mode $n$, we add the term $\mathcal{S}_{n} \times_n \mathbf{U}^{(n)}$ to the edit tensor, where all elements of the mode-$n$ unfolding of $\mathcal{S}_n$ are zero except for its $i^\text{th}$ row that is set to ${\mathbf{S}_n}_{(n)}(i, :) := \alpha_i \cdot \mathds{1}$. We will experimentally show in \cref{sec:exp:modewise} that, broadly speaking, edits along the channel modes generate changes to the style of the image, whilst the major geometric changes can be generated by basis vectors for the spatial modes.

\paragraph{Multilinear mixing}
\label{sec:method:interactions}
In addition to making edits along a single mode, we can alternatively model the multiplicative interactions \cite{jayakumar2019multiplicative} of basis vectors \textit{between} modes--we call this `multilinear mixing'.
To model the third-order interactions of the $i$th, $j$th, and $k$th basis vectors for the three modes respectively, we define a selector tensor $\mathcal{S}_{CHW}\in\mathbb{R}^{C\times H \times W}$
with $\mathcal{S}_{CHW}(i,j,k):=\alpha_{ijk}$, and set the edit tensor as $\mathcal{S}_{CHW} \times_1 \mathbf{U}^{(C)} \times_2 \mathbf{U}^{(H)} \times_3 \mathbf{U}^{(W)}$, that is, a rank-$1$ tensor formed as the outer product $\alpha_{ijk} \cdot \mathbf{u}^{(C)}_i \circ \mathbf{u}^{(H)}_j \circ \mathbf{u}^{(W)}_k$. In the general case when multiple directions are manipulated simultaneously this term can be written as a sum over rank-1 tensors formed by outer products of the desired vectors from each basis, weighted by the elements in $\mathcal{S}_{CHW}$.
Second-order terms are obtained analogously as a rank-1 matrix which is replicated along each slice of the edit tensor with an outer product with the $\mathds{1}$-vector. Combining the \textcolor{red}{$1$st}-, \textcolor{blue}{$2$nd}-, and \textcolor{ForestGreen}{$3$rd}-order terms leads to the most general form of the edit tensor as
\begin{align}\label{eq:interactions}
  \mathcal{Z}' = &\textcolor{red}{\mathcal{S}_C \times_1 \mathbf{U}^{(C)} + \mathcal{S}_H \times_2 \mathbf{U}^{(H)} + \mathcal{S}_W \times_3 \mathbf{U}^{(W)} }\nonumber \\
  + &\textcolor{blue}{\mathcal{S}_{CH} \times_1 \mathbf{U}^{(C)} \times_2 \mathbf{U}^{(H)} + \dots } 
  + \textcolor{ForestGreen}{\mathcal{S}_{CHW} \times_1 \mathbf{U}^{(C)} \times_2 \mathbf{U}^{(H)} \times_3 \mathbf{U}^{(W)}}.
\end{align}

\subsubsection{Mapping edits to latent space}
\label{sec:mapping}

To use these directions found in the activation space to edit the original latent code, we transform this edit tensor back to the original latent space \cite{hark2020ganspace}. Concretely, we seek to learn a mapping from the original activation tensor $\mathcal{Z}\in\mathbb{R}^{C\times H \times W}$ to its corresponding latent code $\mathbf{z}\in\mathbb{R}^{d}$. Using this mapping, we can then generate the latent code $\mathbf{z}'$ for a desired edit tensor $\mathcal{Z}'$.
To solve this, we propose a tensor regression \cite{kossaifi2020tensor,guo2012tensor} of the form
\begin{align}
\mathbf{z} &= \langle \mathcal{Z}, \mathcal{W} \rangle_3 = \sum_{c} \sum_{h} \sum_{w} \mathcal{Z}(c,h,w) \mathcal{W}(c,h,w,:),
\end{align} with weight tensor $\mathcal{W}\in\mathbb{R}^{C\times H \times W \times d}$. That is to say, we take a ``generalised inner product'' \cite{kossaifi2020tensor} along the last $3$ modes of $\mathcal{Z}$ and first $3$ modes of $\mathcal{W}$.
This gives us the objective function for the regression task as
\begin{align}\label{eq:reg}
    \mathcal{L}_{rec} = \mathbb{E}_{\mathbf{z}\sim\mathcal{N}\left(\mathbf{0},\mathbf{I}_d\right)}\left[
    \left|\left|
    \mathbf{z} - \left\langle \mathcal{Z}, \mathcal{W} \right\rangle_3
    \right|\right|_2^2
    \right] + \lambda \left|\left|\mathcal{W} \right|\right|_2^2.
\end{align}
We explore imposing a low-rank Tucker structure on $\mathcal{W}$ in order to reduce the number of parameters and perform regularisation, training the factors of the Tucker decomposition forming $\mathcal{W}$ in \cref{eq:reg} with gradient descent using TensorLy \cite{tensorly}.
Finally, following the linear case in \cref{sec:linear-solution}, we perform the edit with $\mathcal{X}' = G \big( \mathbf{z} + \langle \mathcal{Z}', \mathcal{W} \rangle_3 \big)$, where $\mathcal{Z}'$ is given by \cref{eq:interactions}.

\section{Experiments}
\label{sec:experiments}

In this section, we present a series of experiments to validate the proposed method.
We first showcase in \cref{sec:exp:modewise} how we can find directions along each mode that correspond intuitively to either geometry- or style-based attributes, along with the multilinear mixes.
Finally, we present quantitative results in \cref{sec:exp:quant} that show superior disentanglement compared to the state-of-the-art.

\begin{figure}[h]
    \centering
    \includegraphics[width=1.0\textwidth]{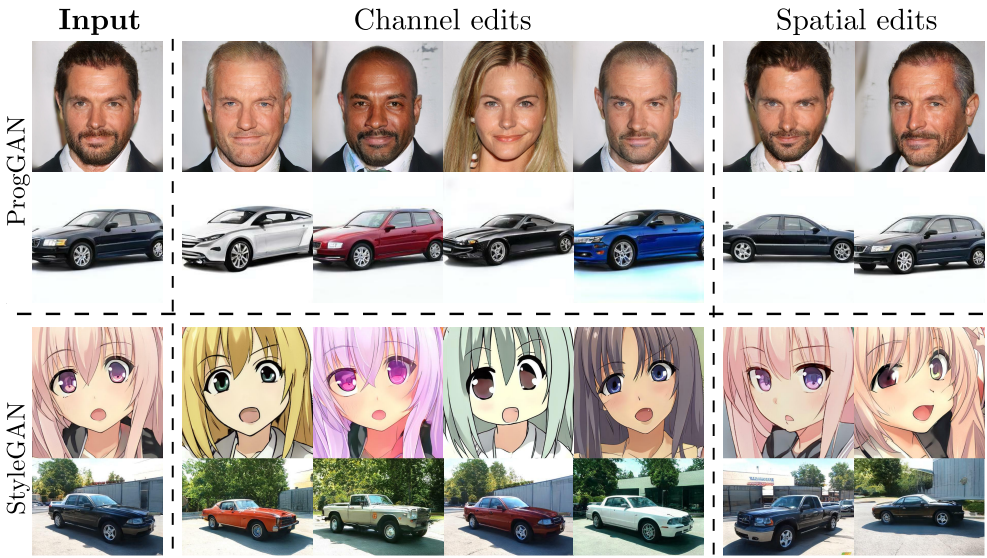}
    \caption{Edits performed along the spatial and channel modes separately, in a variety of generators and datasets. For these experiments, we use a low-rank Tucker decomposition for the regression tensor.}
    \label{fig:single_edits}
\end{figure}

\paragraph{Implementation details}
\label{sec:exp:implementation}

We experiment on both the traditional (transpose) convolutional architecture of \textbf{ProgGAN} and the style-based generator from \textbf{StyleGAN}. We apply the multilinear decomposition after the first block of convolutions for both networks, and map these activations back to their corresponding latent codes (or style vectors, for StyleGAN). The images generated by these networks are not always high quality however, and therefore in this section we manually select initial seeds that produce realistic images to showcase our results on.

\subsection{Qualitative results}
\label{sec:exp:qual}

\paragraph{Mode-wise edits}
\label{sec:exp:modewise}

\begin{figure}[h]
    \centering
    \includegraphics[width=1.0\textwidth]{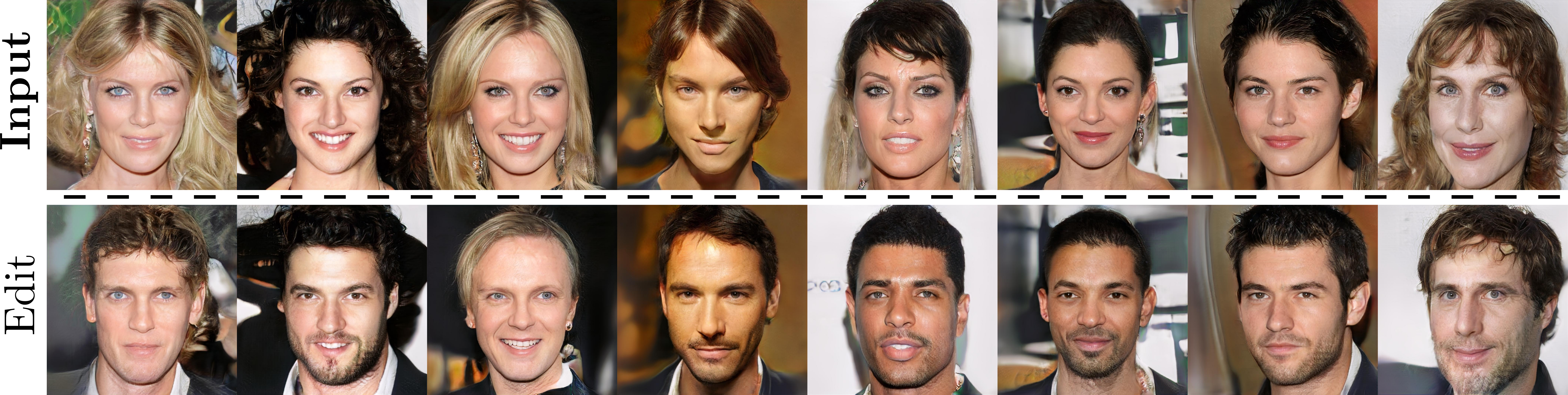}
    \caption{Walking along a single found direction in the channel basis by the same amount for 8 random seed images, we find each image is transformed in the same manner.}
    \label{fig:edits-consistent}
\end{figure}

In this section, we demonstrate qualitatively some of our method's learnt directions for each mode. We find that by making edits along the channels of the activation maps we tend to affect changes to the style of the image, leaving the geometry of the image (such as its pose) largely untouched. This is demonstrated qualitatively in \cref{fig:single_edits}--for example, the channel basis vectors affect semantic changes such as the colour of a person's hair or skin. We show how such directions affect different images in a consistent manner in \cref{fig:edits-consistent}. On the other hand, we frequently find at least one of the major geometry-based directions in the relevant spatial mode. For example, for CelebA-HQ \cite{karras2018prog} on ProgGAN, we find `pitch' in the basis for the horizontal mode, and `yaw' in the vertical basis. As with all PCA-based methods however, the number of semantically relevant directions we can find in each mode is limited by the variation in the original training data.
\begin{figure}[h]
    \centering
    \includegraphics[width=0.8\textwidth]{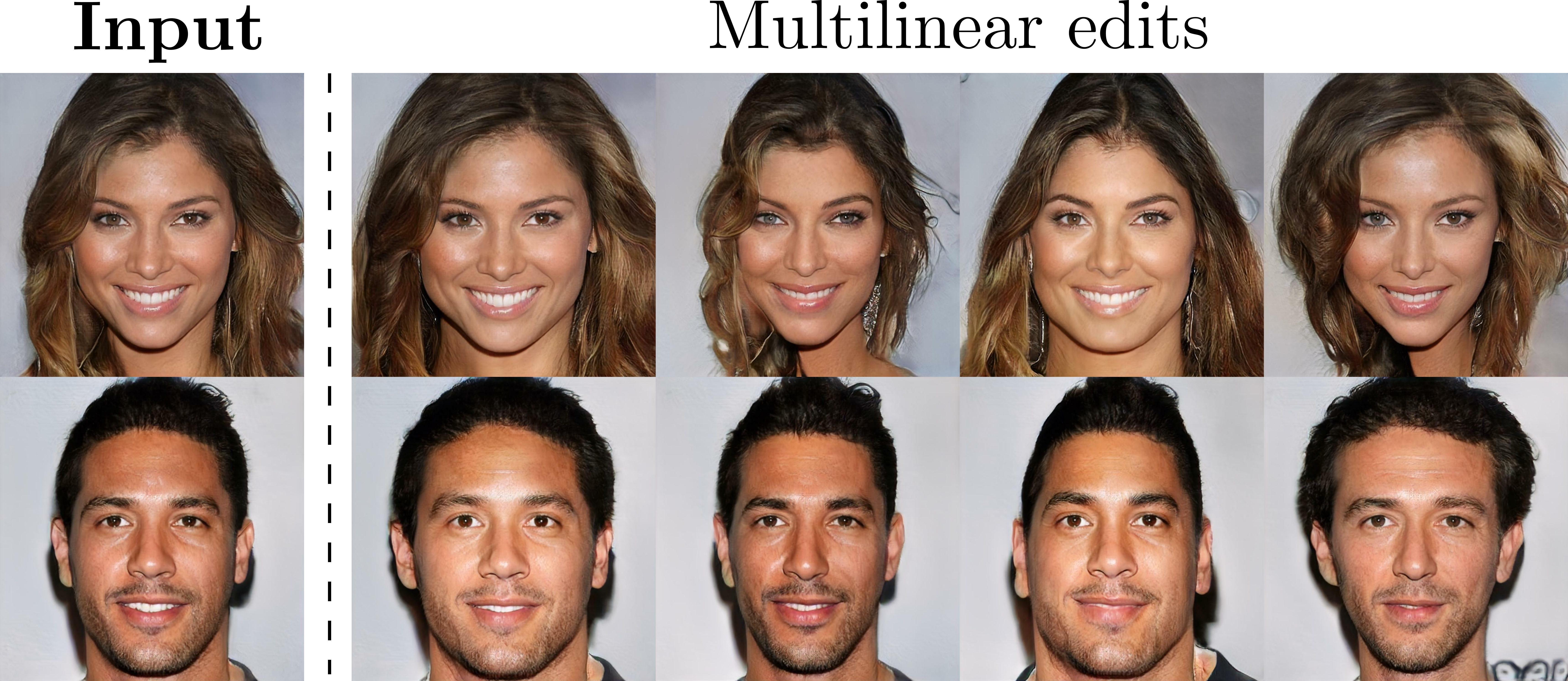}
    \caption{Edits found in the third-order interactions of bases for ProgGAN (with a rank-\texttt{256,4,4,512} Tucker decomposition on the weight tensor).}
    \label{fig:unique-pggan}
\end{figure}

\paragraph{Multilinear mixing}
\label{sec:exp:multilinear}

As detailed in \cref{sec:method:interactions}, our method can alternatively model the interactions of the basis vectors. Here we show experimentally that such `multilinear mixing' can generate a unique set of transformations for which the influence of more than one mode is necessary. In particular, we list examples of these transformations we observe to be possible experimentally in the right-most column of \cref{tab:transformation-types}. We show in \cref{fig:unique-pggan} some of these on ProgGAN, allowing us fine-grained control over attributes such as forehead size and face shape. We find this ability to affect these kinds of coarse changes to not be possible with the baseline methods--highlighting the importance of explicitly treating the modes separately and modelling the interactions between their bases.
We find a high-rank Tucker structure on the weight tensor is beneficial to best capture these coarse changes in the higher-order terms. 

\vspace{1em}

\subsection{Quantitative results}
\label{sec:exp:quant}

\begin{figure}[]
\centering
\subfigure[Unsup. Discov. \cite{voynov2020unsupervised}]{
    \centering
    \includegraphics[width=0.22\textwidth]{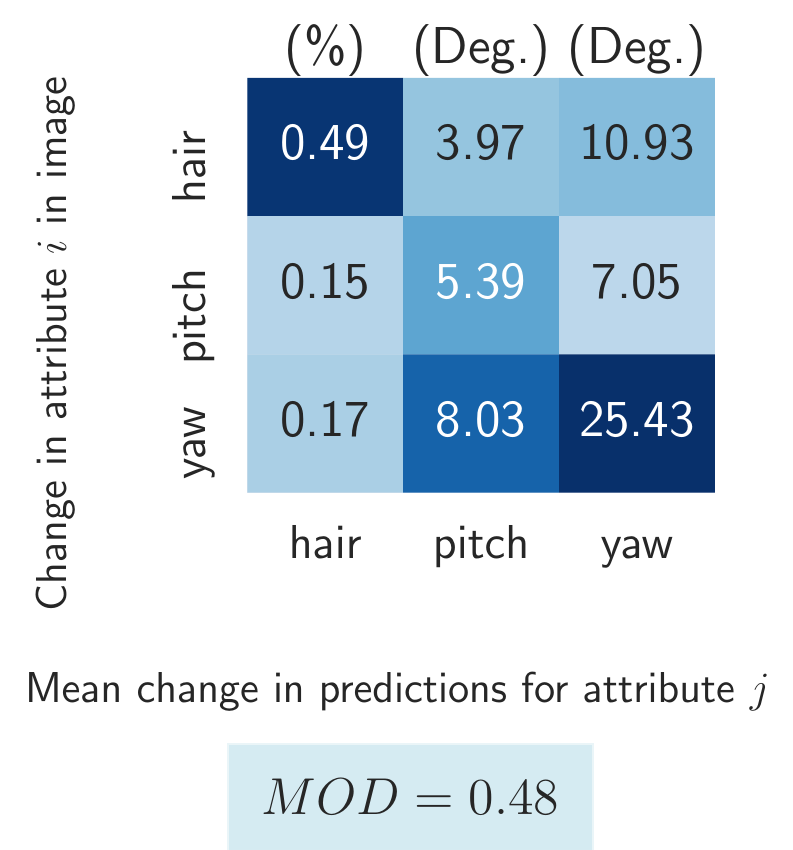}
    \label{fig:dis-quant-quant-babenko}
}
\hfill
\subfigure[SeFa \cite{shen2021closedform}]{
    \centering
    \includegraphics[width=0.22\textwidth]{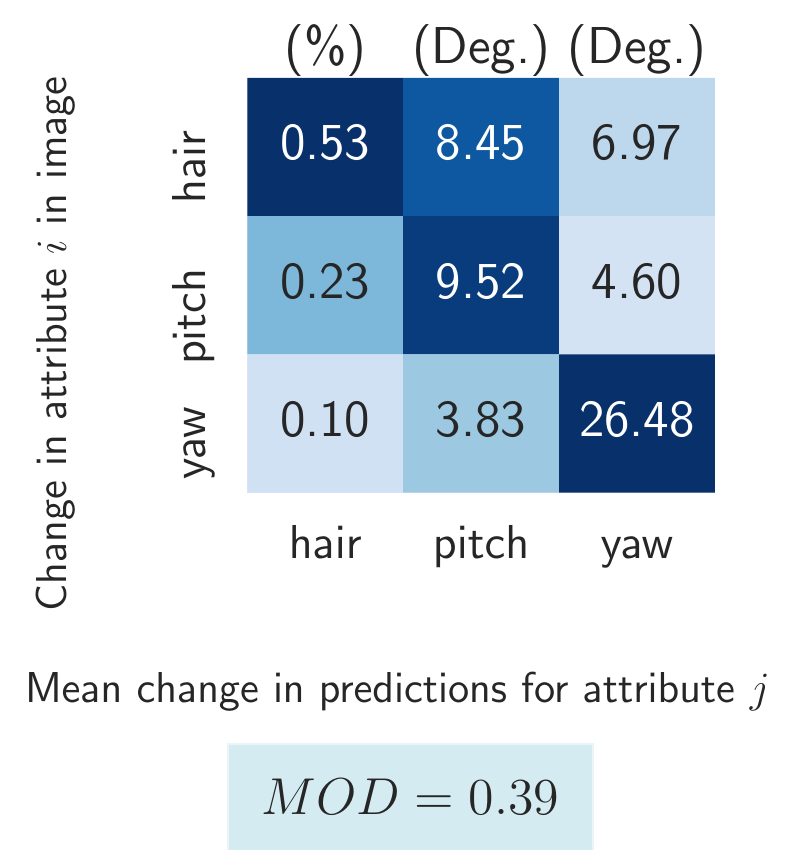}
    \label{fig:dis-quant-quant-sefa}
}
\hfill
\subfigure[GANSpace \cite{hark2020ganspace}]{
    \centering
    \includegraphics[width=0.22\textwidth]{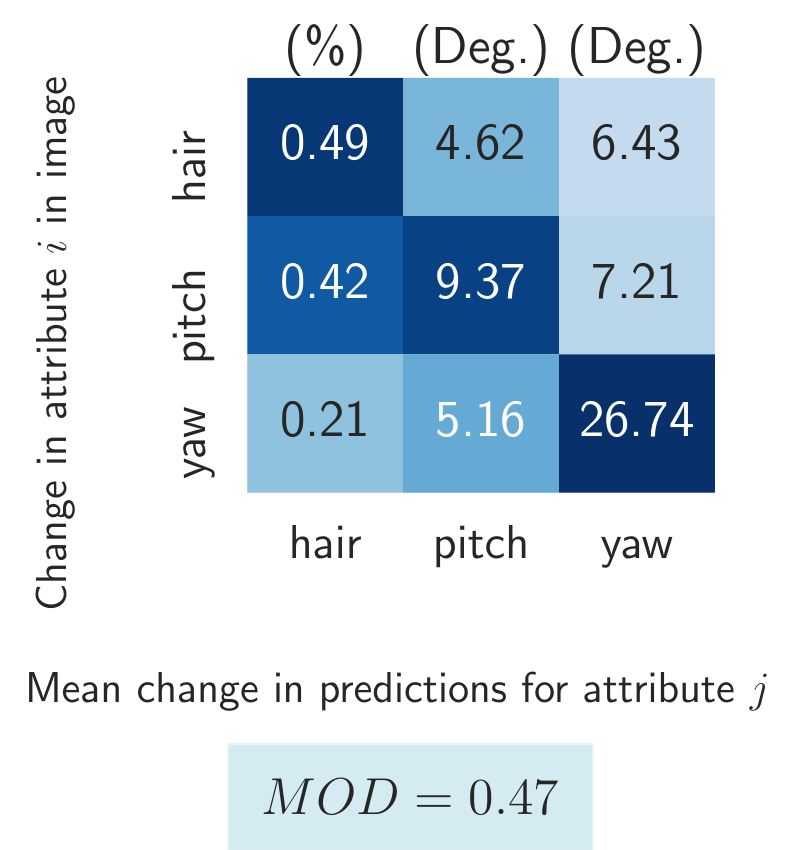}
    \label{fig:dis-quant-quant-gspace}
}
\hfill
\subfigure[\textbf{Multilinear (ours)}]{
    \centering
    \includegraphics[width=0.22\textwidth]{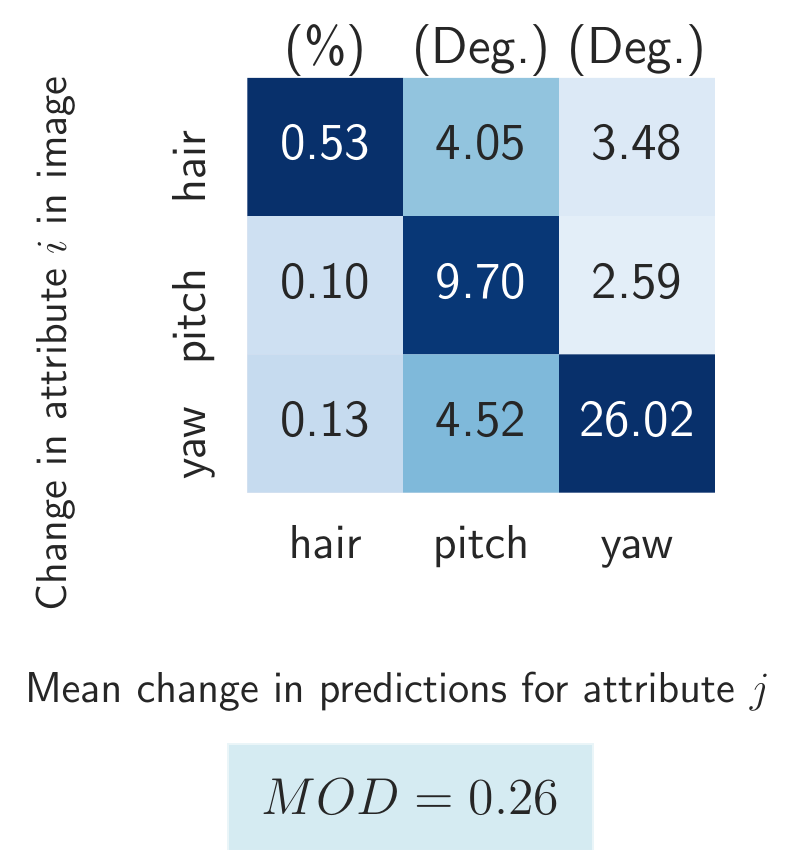}
    \label{fig:dis-quant-quant}
}
  \caption{
  (a) Mean difference in predictions (columns) between the original images and their edited versions for the target attribute (rows), on various baselines. The mean of the (column-normalised) off-diagonals ($MOD$ ($\downarrow$)) is shown below each.}
    \label{fig:dis-quant}
\end{figure}

\begin{figure}[t]
    \centering
    \includegraphics[width=1.0\textwidth]{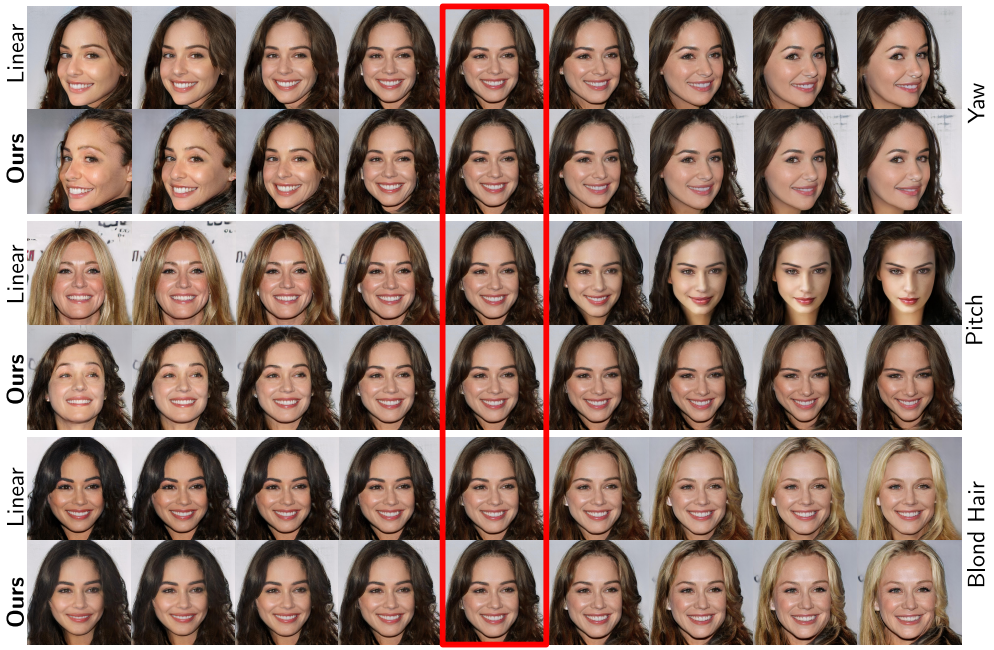}
    \caption{Linearly interpolating along the basis vectors for `yaw', `pitch', and `blond hair' for both our method and GANspace. As can be seen, the linear model's `pitch' direction is entangled with hair colour, whereas ours exhibits clearer visual disentanglement.}
    \label{fig:ganspace-comparison}
\end{figure}

Finally, we quantify how well our method can recover implicitly disentangled directions corresponding to style and geometry.
We use the same model trained on ProgGAN with which we generated \cref{fig:single_edits}, manually identifying a prominent recovered direction for each mode: hair colour (channels), yaw (vertical), and pitch (horizontal). We synthesise 100 random images, and 3 edited versions of each image, walking along the direction corresponding to each of the three attributes. We obtain predictions for these three attributes on all $400$ images using Microsoft Azure\footnote{\url{https://azure.microsoft.com/en-gb/services/cognitive-services/face/}}. Finally, we compute the mean absolute difference between the predictions for each attribute on the unedited images and on the 3 edited versions, to see how changing one attribute affects the predictions for the other two. We collect these values in a matrix $\mathbf{A}\in\mathbb{R}^{N\times N}$ (shown for various baselines in \cref{fig:dis-quant} with $N:=3$), with $a_{i,j}$ being the mean difference in predictions for attribute $i$ between the raw images and those with attribute $j$ edited. It's clear from the small values in the off-diagonals of \cref{fig:dis-quant-quant}, that these directions affect only the single target factor of variation to a greater extent than the baseline results in \cref{fig:dis-quant-quant-babenko,fig:dis-quant-quant-sefa,fig:dis-quant-quant-gspace}. We can quantify this by taking the mean of the off-diagonals ($MOD$) for each method with $MOD=\frac{1}{N^2 - N}\left(\sum_{i,j}\hat{a}_{i,j} - \text{tr}(\mathbf{\hat{A}}) \right)$, where $\mathbf{\hat{A}}$ is the column-normalised $\mathbf{A}$. Our method achieves a notably lower score than all baselines. We see qualitatively in our GANSpace comparison in \cref{fig:ganspace-comparison} that the `pitch' attribute identified in GANspace clearly leaks style information to a greater extent that our direction for the `pitch' attribute.

\section{Conclusion}
\label{sec:conclusion}
In this paper, we have presented a method that offers an intuitive way to find different types of interpretable transformations in a pre-trained GAN. We achieve this by decomposing the generator's activations in a multilinear manner, and regressing back to the latent space.
We showed that one can find directions along each mode of the tensor that correspond semantically to the relevant mode (for example, we find the `yaw' direction in the `vertical' basis). Additionally we showed that by modelling the multiplicative interactions of basis vectors across the modes, we recover an extended set of directions we find to be unobtainable with other methods in the literature. We showed that our found directions achieve superior disentanglement over prior work. In our experiments, we found the choice of rank on the weight tensor to play an important role in the quality of our found directions. In future work we plan to develop techniques to determine the appropriate rank automatically.
    
\paragraph{Acknowledgements} This work was supported by a grant from The Cyprus Institute on Cyclone under project ID p055, and the EU H2020 AI4Media No. 951911 project.

\begin{appendices}
\label{sec:intro}
In this document, we present additional material to support the main paper. Firstly, we provide illustrations and derivations in \cref{sec:intuition}, aimed at clarifying and providing intuition into some of the operations performed in the main paper. Lastly, in \cref{sec:additional-results} we provide experimental results designed to supplement and further validate our proposed method.

\section{Illustrations \& intuition}
\label{sec:intuition}

\subsection{Mode-$n$ edits}
In the main paper, we form an `edit tensor' $\mathcal{Z}'\in\mathbb{R}^{C\times H \times W}$ which is a combination of the basis vectors for each of the three modes of the generator's activations. We show how one can make edits that, broadly speaking, correspond to style or geometry by adding the mode-$n$ basis vectors to all mode-$n$ fibers of this edit tensor, using the $1^\text{st}$ order terms $\mathcal{Z}' = \mathcal{S}_{n} \times_n \mathbf{U}^{(n)}$.

To see how these $1^\text{st}$ order terms work to select the desired linear combinations of the $N$ basis vectors from the columns of $\mathbf{U}^{(n)}$ and sum them along each of the output's mode-$n$ fibers, we can inspect $\mathcal{Z}'$'s mode-$n$ unfolding. We know from the definition of the mode-$n$ (matrix) product \cite{kolda_tensor_2009} that we can write this term equivalently as
\begin{align}
\label{eq:selector-eq}
\mathcal{Z}' = \mathcal{S}_{n} \times_n \mathbf{U}^{(n)} \quad \Leftrightarrow \quad \mathbf{Z}'_{(n)} = \mathbf{U}^{(n)}{\mathbf{S}_n}_{(n)}.
\end{align}
Next, recall that the definition of the mode-$n$ unfolding of a tensor $\mathcal{X}$ is a rearranging of its mode-$n$ fibers into the columns of a matrix $\mathbf{X}_{(n)}$ \cite{kolda_tensor_2009}. With this in mind, we can inspect the right-hand-side of \cref{eq:selector-eq}, writing it as
\begin{align}
    \mathbf{Z}'_{(n)} &= \mathbf{U}^{(n)}{\mathbf{S}_n}_{(n)} \\
    &=
    \underbrace{\begin{bmatrix}
        \vert & &\vert \\
        \mathbf{u}^{(n)}_1 & \cdots & \mathbf{u}^{(n)}_N   \\
        \vert & &\vert
    \end{bmatrix}}_{\mathbf{U}^{(n)}}
    \underbrace{\brows{ \alpha_1\mathds{1}^\top \\ \rowsvdots \\ \alpha_N\mathds{1}^\top }}_{{\mathbf{S}_n}_{(n)}}
    \\
    &= \sum_i \mathbf{u}^{(n)}_i \circ \alpha_i\mathds{1},
\end{align}
which shows that each of the mode-$n$ fibers of $\mathcal{Z}'$ are linear combinations of the mode-$n$ basis vectors, as intended.

\subsection{Multilinear mixing}
We also show we can model the interactions of the basis vectors between the modes of the tensor. We first recall the following useful result with the Kronecker product \cite{kolda2006multilinear}:
\begin{proposition}[]
\label{prop:vec}
    \textit{Let} $\mathcal{Y} = \mathcal{X}
    \times_1 \mathbf{U}^{(1)}
    \times_2 \mathbf{U}^{(2)}
    \times_3 \cdots \times_N \mathbf{U}^{(N)}$\textit{, then}
    \begin{align}
        \text{vec}\left( \mathcal{Y} \right) =
        \left(\mathbf{U}^{(N)} \otimes
        \mathbf{U}^{(N-1)} \otimes \cdots \otimes
        \mathbf{U}^{(1)} \right) \text{vec}\left( \mathcal{X} \right).
    \end{align}
\end{proposition}
The $3^\text{rd}$ order term $\mathcal{Z}' = \mathcal{S}_{CHW} \times_1 \mathbf{U}^{(C)} \times_2 \mathbf{U}^{(H)} \times_3 \mathbf{U}^{(W)}$ can then be understood most easily by appealing to \cref{prop:vec} and writing it in terms of its vectorisation as
\begin{align}
\text{vec} \left( \mathcal{Z}' \right)
    &= \left( \mathbf{U}^{(W)} \otimes \mathbf{U}^{(H)} \otimes \mathbf{U}^{(C)} \right) \text{vec}\left( \mathcal{S}_{CHW} \right) \\
    &= \left( \mathbf{U}^{(W)} \otimes \mathbf{U}^{(H)} \otimes \mathbf{U}^{(C)} \right)
    \underbrace{\begin{bmatrix}
        \alpha_1 \\
        \alpha_2 \\
        \vdots \\
        \alpha_{C\cdot H \cdot W}
    \end{bmatrix}}_{\text{vec}\left( \mathcal{S}_{CHW} \right)}.
\end{align}
That is, considering the operation in its vectorised form, the `selector tensor' $\mathcal{S}_{CHW}$ can be interpreted as simply taking a linear combination of the columns of the matrix formed from the interactions of the basis vectors of the three factor matrices. For example, $\text{vec}\left( \mathcal{S}_{CHW} \right)(1):= \alpha_1$ weights the interactions of the first basis vectors of all three of the bases $\mathbf{u}^{(C)}_1, \mathbf{u}^{(H)}_1, \mathbf{u}^{(W)}_1$.

\subsection{Regressing the edit tensor to the latent code}
\begin{figure}[h]
    \centering
    \includegraphics[width=1.0\linewidth]{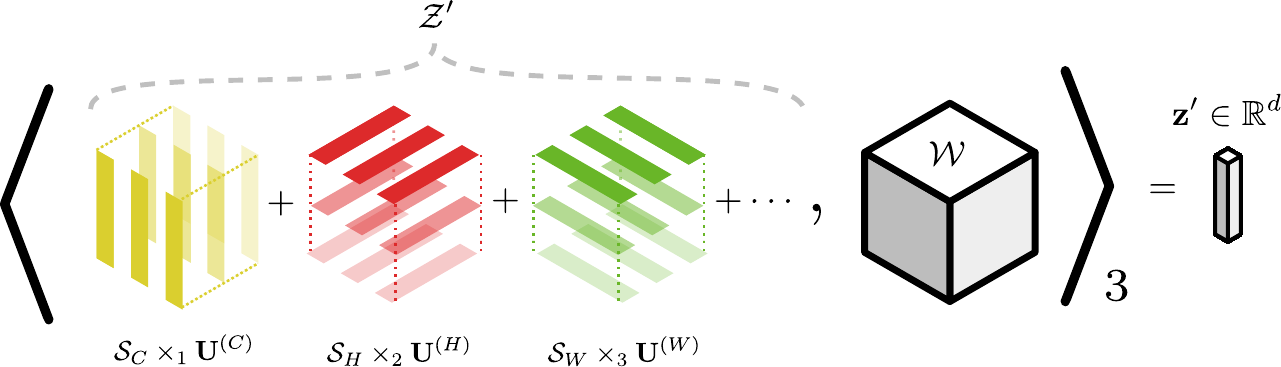}
    \caption{An overview of how we form the edit tensor and compute its latent code: We first form the edit tensor $\mathcal{Z}'$, the $1^\text{st}$-order terms being demonstrated graphically here. Then, we take the generalised inner product with weight tensor $\mathcal{W}$ yielding $\mathbf{z}'$: the corresponding direction in the original latent space.}
    \label{fig:regression}
\end{figure}

Finally, we illustrate how these terms are summed to form the edit tensor graphically in \cref{fig:regression}. The `generalised inner product' then maps this back to the latent code $\mathbf{z}'\in\mathbb{R}^d$.

\subsection{Computing the bases in a pre-trained GAN}

To compute the three bases $\mathbf{U}^{(C)},\mathbf{U}^{(H)},\mathbf{U}^{(W)}$ we follow \cite{haiping_lu_mpca_2008}, where Lu et al. show that, if we retain each of the basis vectors (i.e., perform no dimensionality reduction), we can compute these factor matrices in one-shot. Given a pre-trained GAN's generator $G$, we compute the bases following \cref{algo:bases}.
\begin{algorithm}[H]
\caption{Computing the (full rank) multilinear bases}
\label{algo:bases}
\begin{algorithmic}[1]
\Procedure{ComputeBases}{$G,i$}\Comment{Pretrained generator $G$ and target layer $i$}
 \State $\mathbf{Z} \gets \text{Sample } M \text{ times from standard normal}$
 \State $\mathcal{Z} \gets G[:i](\mathbf{Z})\in\mathbb{R}^{M\times C\times H \times W}$\Comment{Intermediate activations}
\For{$n=1:3$}
    \State $\mathbf{\bar{Z}}_{(n)} \gets \frac{1}{M} \sum_{m=1}^M \mathbf{Z}_{m(n)}$\Comment{Mean mode-$n$ unfoldings}
    \State $\mathbf{U}^{(n)} \gets \text{Left-singular vectors of } \sum_{m=1}^M \left(\mathbf{Z}_{m(n)} - \mathbf{\bar{Z}}_{(n)}\right) \left(\mathbf{Z}_{m(n)} - \mathbf{\bar{Z}}_{(n)}\right)^\top$
\EndFor
\State \textbf{return} $\mathbf{U}^{(1)},\mathbf{U}^{(2)},\mathbf{U}^{(3)}$
\EndProcedure
\end{algorithmic}
\end{algorithm}

\section{Additional experimental results}
\label{sec:additional-results}

\subsection{Qualitative results}

\begin{figure}[t]
    \centering
    \includegraphics[width=1.0\textwidth]{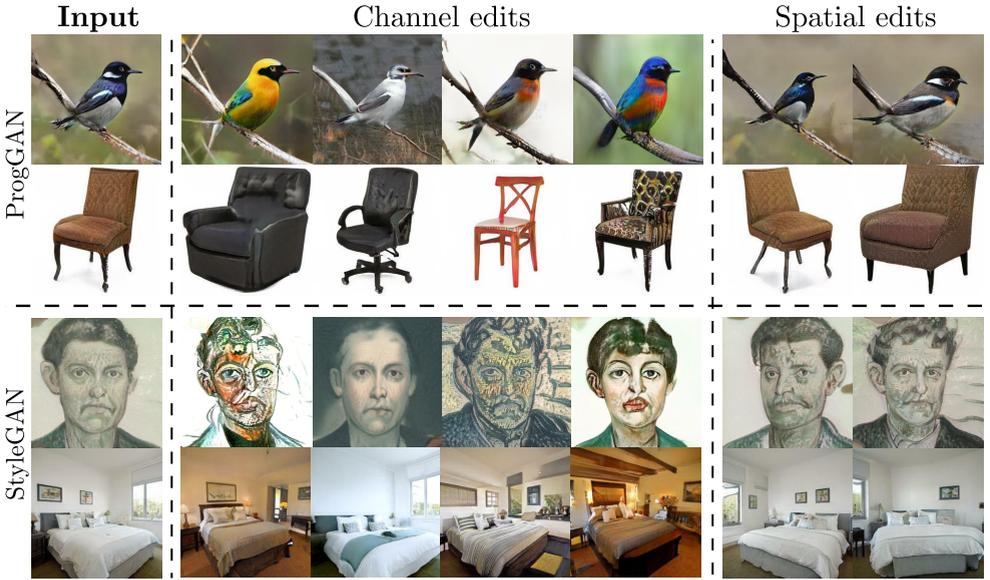}
    \caption{Edits performed along the spatial and channel modes separately, in a variety of generators and datasets. For these experiments, we use a low-rank Tucker decomposition for the regression tensor.}
    \label{fig:single_edits_sup}
\end{figure}

\begin{figure}
    \centering
    \includegraphics[width=1.0\textwidth]{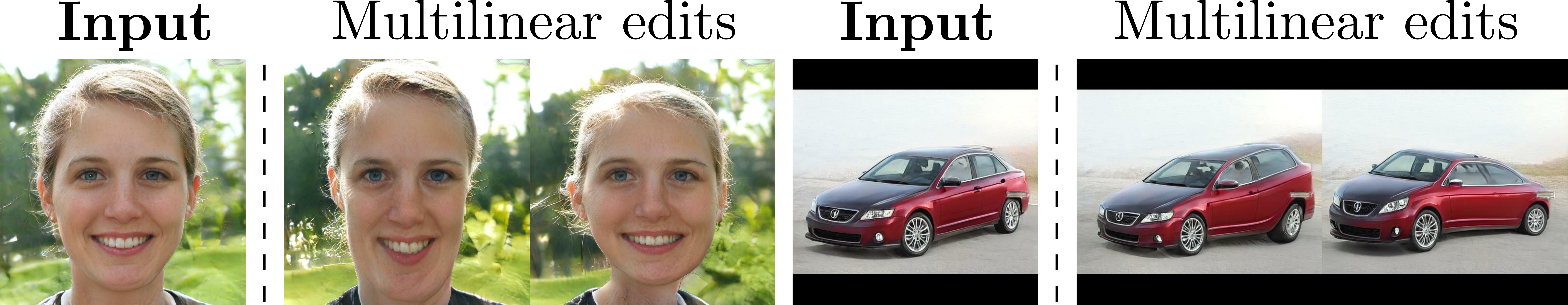}
    \caption{Edits found in the third-order interactions of bases for StyleGAN (when we directly edit the activation tensor). }
    \label{fig:unique-sgan}
\end{figure}

Here we provide additional qualitative results. We first show more mode-wise edits in \cref{fig:single_edits_sup}. In \cref{fig:unique-sgan} we also perform `multilinear mixing' on StyleGAN. For StyleGAN, we find these multilinear directions are best realised when we additionally add the edit tensor to the activation tensor directly--along with generating the corresponding latent code for the AdaIN operations--and continue the forward pass from this edited tensor.

\subsection{Experimental setup}
Here we describe more thoroughly the experimental setup we use to produce our results in the quantitative comparisons in the main paper, to ensure reproducibility. In general, we walk along each direction by a manual amount for each direction of each baseline, such that the total average mean change in predictions for each attribute is as close as possible. E.g. we walk along the `blonde hair' direction for all methods until the mean change in this attribute is close to $0.5$. The baselines we benchmark our proposed method against are detailed below.

\paragraph{GANSpace} For GANSpace \cite{hark2020ganspace}, no official weights are provided for the CelebA-HQ dataset for the ProgGAN generator. Therefore we manually implement this on top of the author's official code\footnote{GANSpace codebase: \url{https://github.com/harskish/ganspace}}, using a total of $1,000,000$ samples to perform the decomposition.

\paragraph{SeFA} For SeFA \cite{shen2021closedform}, we use the author's official code and pre-trained weights\footnote{Official SeFA weights: \url{https://github.com/genforce/sefa/blob/master/latent_codes/pggan_celebahq1024_latents.npy}} to produce the edited images, manually identifying the directions that most closely correspond to the three attributes of interest. Concretely, we use indices \texttt{2,4,7} of the directions matrix for the attributes `yaw', `blond hair', and `pitch' respectively.

\paragraph{Unsupervised discovery} For \cite{voynov2020unsupervised}, the official weights provided are trained on a different ProgGAN model to the other baselines. For this reason, we generate and use the predictions on a new ground-truth training set from this pre-trained model for this baseline for fair comparison. Using the pre-trained weights, we use indices \texttt{5,12,49} of the directions matrix for the attributes `blond hair', `yaw', and `pitch' respectively.

\paragraph{Ours} For our method, we use the indices \texttt{3,3,1} of the channel, height, and width bases for the attributes `blond hair', `pitch', and `yaw' directions respectively.

\subsection{Ablation study: choice of rank for $\mathcal{W}$}

\begin{figure}[h]
    \centering
    \includegraphics[width=0.5\textwidth]{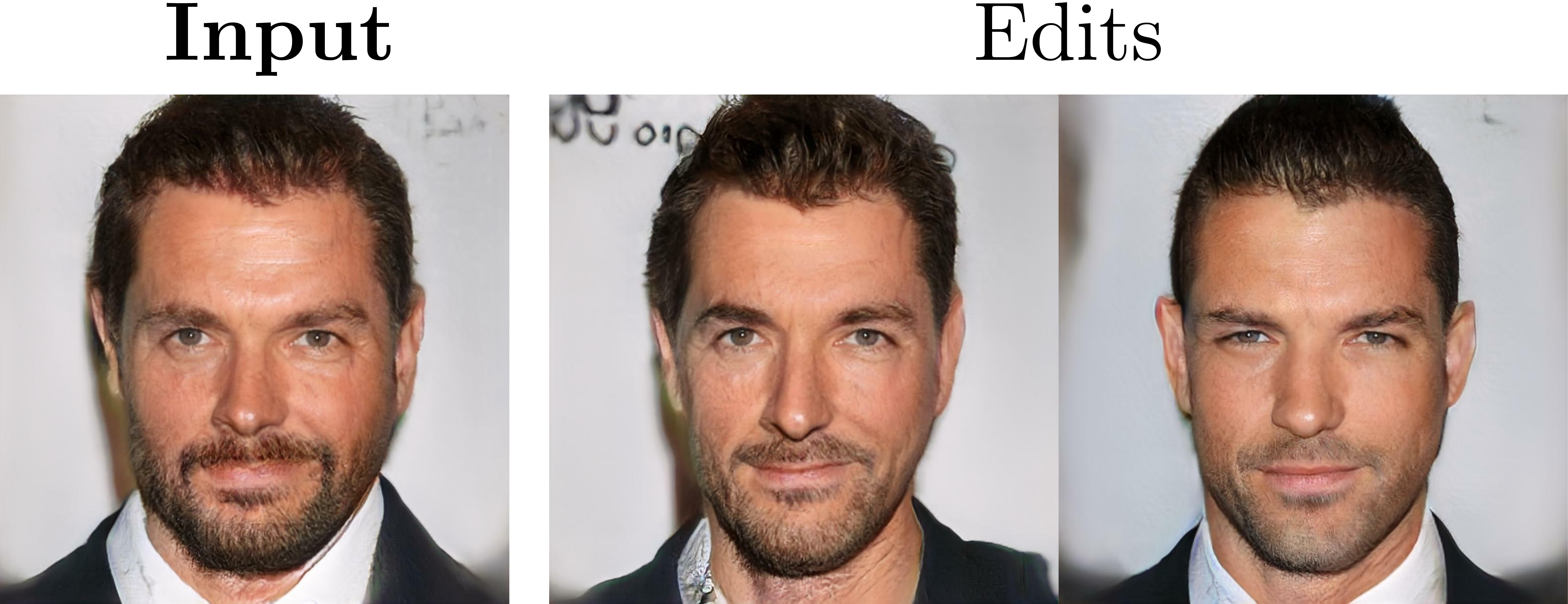}
    \caption{An example of a multilinear direction affecting the head width, generated using both a high- (a) and low-rank (b) regression tensor.}
    \label{fig:ablation}
\end{figure}

We briefly turn our focus to exploring the role of the regression tensor $\mathcal{W}$. We find the regularisation afforded by the choice of rank in the decomposition in regression tensor $\mathcal{W}$ to play an important role in the kind of edits we can generate. For example, we show in \cref{fig:ablation} the head-thinning multilinear direction using both a high- and low-rank Tucker structure on the regression tensor. We find a high rank necessary to generate the `multilinear mixing' edits (that feature transformations very far from the true data distribution). However for smooth, interpretable directions such as pitch and yaw, we find a low-rank necessary--we find a high-rank can lead to artefacts for the first-order terms. Our findings here suggest that the rank of the decomposition of the tensor regression weight is a hyperparameter that one can tune depending on the types of edits they wish to generate. In practice we regress these terms back to the latent code separately.

\end{appendices}

\bibliography{main}

\begin{thebibliography}{40}
\providecommand{\natexlab}[1]{#1}
\providecommand{\url}[1]{\texttt{#1}}
\expandafter\ifx\csname urlstyle\endcsname\relax
  \providecommand{\doi}[1]{doi: #1}\else
  \providecommand{\doi}{doi: \begingroup \urlstyle{rm}\Url}\fi

\bibitem[Broad et~al.(2021)Broad, Leymarie, and Grierson]{broad2020bending}
Terence Broad, Frederic~Fol Leymarie, and Mick Grierson.
\newblock Network bending: Expressive manipulation of deep generative models.
\newblock In \emph{Artificial Intelligence in Music, Sound, Art and Design},
  pages 20--36. Springer International Publishing, 2021.

\bibitem[Brock et~al.(2019)Brock, Donahue, and Simonyan]{brock2018large}
Andrew Brock, Jeff Donahue, and Karen Simonyan.
\newblock Large scale {GAN} training for high fidelity natural image synthesis.
\newblock In \emph{Proc. Int. Conf. Mach. Learn. (ICML)}, 2019.

\bibitem[Bulat et~al.(2020)Bulat, Kossaifi, Tzimiropoulos, and
  Pantic]{bulat2020incremental}
Adrian Bulat, Jean Kossaifi, Georgios Tzimiropoulos, and Maja Pantic.
\newblock Incremental multi-domain learning with network latent tensor
  factorization.
\newblock In \emph{Proc. AAAI Conf. Artif. Intell. (AAAI)}, volume~34, pages
  10470--10477, 2020.

\bibitem[Chen et~al.(2016)Chen, Duan, Houthooft, Schulman, Sutskever, and
  Abbeel]{infogan}
Xi~Chen, Yan Duan, Rein Houthooft, John Schulman, Ilya Sutskever, and Pieter
  Abbeel.
\newblock Info{GAN}: Interpretable representation learning by information
  maximizing generative adversarial nets.
\newblock In \emph{Proc. Adv. Neural Inform. Process. Syst. (NeurIPS)},
  volume~29, 2016.

\bibitem[Chrysos et~al.(2020)Chrysos, Moschoglou, Bouritsas, Panagakis, Deng,
  and Zafeiriou]{chrysos2020p}
Grigorios~G Chrysos, Stylianos Moschoglou, Giorgos Bouritsas, Yannis Panagakis,
  Jiankang Deng, and Stefanos Zafeiriou.
\newblock P-nets: Deep polynomial neural networks.
\newblock In \emph{Proc. IEEE Conf. Comput. Vis. Pattern Recog. (CVPR)}, pages
  7325--7335, 2020.

\bibitem[Georgopoulos et~al.(2020)Georgopoulos, Chrysos, Pantic, and
  Panagakis]{multilinear2020georgopoulos}
Markos Georgopoulos, Grigorios Chrysos, Maja Pantic, and Yannis Panagakis.
\newblock Multilinear latent conditioning for generating unseen attribute
  combinations.
\newblock In \emph{Proc. Int. Conf. Mach. Learn. (ICML)}, volume 119, pages
  3442--3451, 2020.

\bibitem[Georgopoulos et~al.(2021)Georgopoulos, Oldfield, Nicolaou, Panagakis,
  and Pantic]{mitigating2021georgopoulos}
Markos Georgopoulos, James Oldfield, Mihalis Nicolaou, Yannis Panagakis, and
  Maja Pantic.
\newblock Mitigating demographic bias in facial datasets with style-based
  multi-attribute transfer.
\newblock \emph{Int. J. Comput. Vis. (IJCV)}, 2021.

\bibitem[Goetschalckx et~al.(2019)Goetschalckx, Andonian, Oliva, and
  Isola]{goetschalckx2019ganalyze}
Lore Goetschalckx, Alex Andonian, Aude Oliva, and Phillip Isola.
\newblock Ganalyze: Toward visual definitions of cognitive image properties.
\newblock In \emph{Proc. Int. Conf. Comput. Vis. (ICCV)}, pages 5744--5753,
  2019.

\bibitem[Goodfellow et~al.(2014)Goodfellow, Pouget-Abadie, Mirza, Xu,
  Warde-Farley, Ozair, Courville, and Bengio]{goodfellow2014gan}
Ian Goodfellow, Jean Pouget-Abadie, Mehdi Mirza, Bing Xu, David Warde-Farley,
  Sherjil Ozair, Aaron Courville, and Yoshua Bengio.
\newblock Generative adversarial nets.
\newblock In \emph{Proc. Adv. Neural Inform. Process. Syst. (NeurIPS)}, 2014.

\bibitem[{Guo} et~al.(2012){Guo}, {Kotsia}, and {Patras}]{guo2012tensor}
W.~{Guo}, I.~{Kotsia}, and I.~{Patras}.
\newblock Tensor learning for regression.
\newblock \emph{IEEE Trans. Image Process.}, 21\penalty0 (2):\penalty0
  816--827, 2012.

\bibitem[{Haiping Lu} et~al.(2008){Haiping Lu}, Plataniotis, and
  Venetsanopoulos]{haiping_lu_mpca_2008}
{Haiping Lu}, K.N. Plataniotis, and A.N. Venetsanopoulos.
\newblock {MPCA}: {Multilinear} {Principal} {Component} {Analysis} of {Tensor}
  {Objects}.
\newblock \emph{{IEEE} Trans. Neural Netw.}, 19:\penalty0 18--39, 2008.

\bibitem[H\"{a}rk\"{o}nen et~al.(2020)H\"{a}rk\"{o}nen, Hertzmann, Lehtinen,
  and Paris]{hark2020ganspace}
Erik H\"{a}rk\"{o}nen, Aaron Hertzmann, Jaakko Lehtinen, and Sylvain Paris.
\newblock {GANS}pace: Discovering interpretable {GAN} controls.
\newblock In \emph{Proc. Adv. Neural Inform. Process. Syst. (NeurIPS)},
  volume~33, pages 9841--9850, 2020.

\bibitem[Huang and Belongie(2017)]{huang2017arbitrary}
Xun Huang and Serge Belongie.
\newblock Arbitrary style transfer in real-time with adaptive instance
  normalization.
\newblock In \emph{Proc. Int. Conf. Comput. Vis. (ICCV)}, pages 1501--1510,
  2017.

\bibitem[Jahanian et~al.(2020)Jahanian, Chai, and Isola]{gansteerability}
Ali Jahanian, Lucy Chai, and Phillip Isola.
\newblock On the "steerability" of generative adversarial networks.
\newblock In \emph{Proc. Int. Conf. Learn. Represent. (ICLR)}, 2020.

\bibitem[Jayakumar et~al.(2019)Jayakumar, Czarnecki, Menick, Schwarz, Rae,
  Osindero, Teh, Harley, and Pascanu]{jayakumar2019multiplicative}
Siddhant~M Jayakumar, Wojciech~M Czarnecki, Jacob Menick, Jonathan Schwarz,
  Jack Rae, Simon Osindero, Yee~Whye Teh, Tim Harley, and Razvan Pascanu.
\newblock Multiplicative interactions and where to find them.
\newblock In \emph{Proc. Int. Conf. Learn. Represent. (ICLR)}, 2019.

\bibitem[Karras et~al.(2018)Karras, Aila, Laine, and Lehtinen]{karras2018prog}
Tero Karras, Timo Aila, Samuli Laine, and Jaakko Lehtinen.
\newblock Progressive growing of {GAN}s for improved quality, stability, and
  variation.
\newblock In \emph{Proc. Int. Conf. Learn. Represent. (ICLR)}, 2018.

\bibitem[Karras et~al.(2019)Karras, Laine, and Aila]{karras2019style}
Tero Karras, Samuli Laine, and Timo Aila.
\newblock A style-based generator architecture for generative adversarial
  networks.
\newblock In \emph{Proc. IEEE Conf. Comput. Vis. Pattern Recog. (CVPR)}, June
  2019.

\bibitem[Karras et~al.(2020)Karras, Laine, Aittala, Hellsten, Lehtinen, and
  Aila]{Karras_2020_CVPR}
Tero Karras, Samuli Laine, Miika Aittala, Janne Hellsten, Jaakko Lehtinen, and
  Timo Aila.
\newblock Analyzing and improving the image quality of {S}tyle{GAN}.
\newblock In \emph{Proc. IEEE Conf. Comput. Vis. Pattern Recog. (CVPR)}, June
  2020.

\bibitem[Kolda(2006)]{kolda2006multilinear}
Tamara~G. Kolda.
\newblock Multilinear operators for higher-order decompositions.
\newblock Technical report, Sandia National Laboratories, April 2006.

\bibitem[Kolda and Bader(2009)]{kolda_tensor_2009}
Tamara~G. Kolda and Brett~W. Bader.
\newblock Tensor {Decompositions} and {Applications}.
\newblock \emph{SIAM Review}, 51\penalty0 (3):\penalty0 455--500, August 2009.

\bibitem[Kossaifi et~al.(2019)Kossaifi, Panagakis, Anandkumar, and
  Pantic]{tensorly}
Jean Kossaifi, Yannis Panagakis, Anima Anandkumar, and Maja Pantic.
\newblock Tensor{Ly}: Tensor learning in {P}ython.
\newblock \emph{J. Mach. Learn. Res.}, 20\penalty0 (26), 2019.

\bibitem[Kossaifi et~al.(2020{\natexlab{a}})Kossaifi, Lipton, Kolbeinsson,
  Khanna, Furlanello, and Anandkumar]{kossaifi2020tensor}
Jean Kossaifi, Zachary~C Lipton, Arinbj{\"o}rn Kolbeinsson, Aran Khanna,
  Tommaso Furlanello, and Anima Anandkumar.
\newblock Tensor regression networks.
\newblock \emph{J. Mach. Learn. Res.}, 21:\penalty0 1--21, 2020{\natexlab{a}}.

\bibitem[Kossaifi et~al.(2020{\natexlab{b}})Kossaifi, Toisoul, Bulat,
  Panagakis, Hospedales, and Pantic]{kossaifi2020factorized}
Jean Kossaifi, Antoine Toisoul, Adrian Bulat, Yannis Panagakis, Timothy~M
  Hospedales, and Maja Pantic.
\newblock Factorized higher-order {CNN}s with an application to spatio-temporal
  emotion estimation.
\newblock In \emph{Proc. IEEE Conf. Comput. Vis. Pattern Recog. (CVPR)}, pages
  6060--6069, 2020{\natexlab{b}}.

\bibitem[Lee et~al.(2020)Lee, Kim, Hong, and Lee]{highfid2020wonkwang}
Wonkwang Lee, Donggyun Kim, Seunghoon Hong, and Honglak Lee.
\newblock High-fidelity synthesis with disentangled representation.
\newblock In \emph{Proc. Eur. Conf. Comput. Vis. (ECCV)}, pages 157--174, 2020.

\bibitem[Lu et~al.(2011)Lu, Plataniotis, and
  Venetsanopoulos]{lu2011subspacesurvey}
Haiping Lu, Konstantinos Plataniotis, and Anastasios Venetsanopoulos.
\newblock A survey of multilinear subspace learning for tensor data.
\newblock \emph{Pattern Recog.}, 44:\penalty0 1540--1551, 08 2011.

\bibitem[Lu et~al.(2013)Lu, Plataniotis, and
  Venetsanopoulos]{lu2013subspacebook}
Haiping Lu, Konstantinos~N. Plataniotis, and Anastasios Venetsanopoulos.
\newblock \emph{Multilinear Subspace Learning: Dimensionality Reduction of
  Multidimensional Data}.
\newblock Chapman \& Hall/CRC, 1st edition, 2013.

\bibitem[Panagakis et~al.(2021)Panagakis, Kossaifi, Chrysos, Oldfield,
  Nicolaou, Anandkumar, and Zafeiriou]{tensormethods2021panagakis}
Yannis Panagakis, Jean Kossaifi, Grigorios~G. Chrysos, James Oldfield,
  Mihalis~A. Nicolaou, Anima Anandkumar, and Stefanos Zafeiriou.
\newblock Tensor methods in computer vision and deep learning.
\newblock \emph{Proc. IEEE Proc.}, 109\penalty0 (5):\penalty0 863--890, 2021.

\bibitem[Plumerault et~al.(2020)Plumerault, {Le Borgne}, and
  Hudelot]{plumerault20iclr}
Antoine Plumerault, Herv{\'e} {Le Borgne}, and C{\'e}line Hudelot.
\newblock Controlling generative models with continuous factors of variations.
\newblock In \emph{Proc. Int. Conf. Learn. Represent. (ICLR)}, 2020.

\bibitem[Radford et~al.(2016)Radford, Metz, and Chintala]{radford2016dcgan}
Alec Radford, Luke Metz, and Soumith Chintala.
\newblock Unsupervised representation learning with deep convolutional
  generative adversarial networks.
\newblock In \emph{Proc. Int. Conf. Learn. Represent. (ICLR)}, 2016.

\bibitem[Shen and Zhou(2021)]{shen2021closedform}
Yujun Shen and Bolei Zhou.
\newblock Closed-form factorization of latent semantics in {GANs}.
\newblock In \emph{Proc. IEEE Conf. Comput. Vis. Pattern Recog. (CVPR)}, 2021.

\bibitem[Shen et~al.(2020{\natexlab{a}})Shen, Gu, Tang, and
  Zhou]{shen2020interpreting}
Yujun Shen, Jinjin Gu, Xiaoou Tang, and Bolei Zhou.
\newblock Interpreting the latent space of {GAN}s for semantic face editing.
\newblock In \emph{Proc. IEEE Conf. Comput. Vis. Pattern Recog. (CVPR)},
  2020{\natexlab{a}}.

\bibitem[Shen et~al.(2020{\natexlab{b}})Shen, Yang, Tang, and
  Zhou]{shen2020interfacegan}
Yujun Shen, Ceyuan Yang, Xiaoou Tang, and Bolei Zhou.
\newblock Inter{F}ace{GAN}: Interpreting the disentangled face representation
  learned by {GANs}.
\newblock \emph{IEEE Trans. Pattern Anal. Mach. Intell.}, 2020{\natexlab{b}}.

\bibitem[Tenenbaum and Freeman(2000)]{tenenbaum_separating_2000}
Joshua~B. Tenenbaum and William~T. Freeman.
\newblock Separating {Style} and {Content} with {Bilinear} {Models}.
\newblock \emph{Neural Computation}, 12\penalty0 (6):\penalty0 1247--1283, June
  2000.

\bibitem[Tzelepis et~al.(2021)Tzelepis, Tzimiropoulos, and
  Patras]{Tzelepis_2021_ICCV}
Christos Tzelepis, Georgios Tzimiropoulos, and Ioannis Patras.
\newblock {WarpedGANSpace}: Finding non-linear {RBF} paths in {GAN} latent
  space.
\newblock In \emph{Proc. Int. Conf. Comput. Vis. (ICCV)}, pages 6393--6402,
  October 2021.

\bibitem[Vasilescu and Terzopoulos(2002)]{vasilescu_multilinear_2002}
M.~Alex~O. Vasilescu and Demetri Terzopoulos.
\newblock Multilinear {Analysis} of {Image} {Ensembles}: {TensorFaces}.
\newblock In \emph{Proc. Eur. Conf. Comput. Vis. (ECCV)}, volume 2350, pages
  447--460. Berlin, Heidelberg, 2002.

\bibitem[Voynov and Babenko(2020)]{voynov2020unsupervised}
Andrey Voynov and Artem Babenko.
\newblock Unsupervised discovery of interpretable directions in the {GAN}
  latent space.
\newblock In \emph{Proc. Int. Conf. Mach. Learn. (ICML)}, pages 9786--9796,
  2020.

\bibitem[Voynov et~al.(2020)Voynov, Morozov, and Babenko]{voynov2020big}
Andrey Voynov, Stanislav Morozov, and Artem Babenko.
\newblock Big {GANs} are watching you: Towards unsupervised object segmentation
  with off-the-shelf generative models.
\newblock \emph{arXiv preprint arXiv:2006.04988}, 2020.

\bibitem[Wang et~al.(2017{\natexlab{a}})Wang, Panagakis, Snape, and
  Zafeiriou]{wang_learning_2017}
Mengjiao Wang, Yannis Panagakis, Patrick Snape, and Stefanos Zafeiriou.
\newblock Learning the {Multilinear} {Structure} of {Visual} {Data}.
\newblock In \emph{Proc. IEEE Conf. Comput. Vis. Pattern Recog. (CVPR)}, pages
  6053--6061, July 2017{\natexlab{a}}.

\bibitem[Wang et~al.(2017{\natexlab{b}})Wang, Panagakis, Snape, and
  Zafeiriou]{wang2017disentangling}
Mengjiao Wang, Yannis Panagakis, Patrick Snape, and Stefanos~P Zafeiriou.
\newblock Disentangling the modes of variation in unlabelled data.
\newblock \emph{IEEE Trans. Pattern Anal. Mach. Intell.}, 40\penalty0
  (11):\penalty0 2682--2695, 2017{\natexlab{b}}.

\bibitem[Wang et~al.(2019)Wang, Shu, Cheng, Panagakis, Samaras, and
  Zafeiriou]{wang2019adversarial}
Mengjiao Wang, Zhixin Shu, Shiyang Cheng, Yannis Panagakis, Dimitris Samaras,
  and Stefanos Zafeiriou.
\newblock An adversarial neuro-tensorial approach for learning disentangled
  representations.
\newblock \emph{Int. J. Comput. Vis. (IJCV)}, 127\penalty0 (6):\penalty0
  743--762, 2019.

\end{thebibliography}

\end{document}